\documentclass[sigconf]{acmart}

\usepackage[linesnumbered,ruled,vlined]{algorithm2e}
\usepackage{amsfonts} 
\usepackage{bbm, dsfont}
\usepackage{multirow}
\usepackage{tikz}
\usepackage{xcolor}
\usepackage{arydshln}
\usepackage{enumitem}
\usepackage{capt-of}
\usepackage{hyperref}
\hypersetup{
    colorlinks=true,
    linkcolor=blue,
    filecolor=magenta,      
    urlcolor=blue
    }

\SetCommentSty{mycommfont}

\newcommand*\circled[1]{\tikz[baseline=(char.base)]{
            \node[shape=circle,fill,inner sep=1pt] (char) {\textcolor{white}{#1}};}}

\newcommand{\semiea}{\textsf{STEA}\text{ }}
\newcommand{\methodname}{\textsf{STEA} }

\AtBeginDocument{%
  }

\copyrightyear{2023} 
\acmYear{2023} 
\setcopyright{acmlicensed}\acmConference[WSDM '23]{Proceedings of the Sixteenth ACM International Conference on Web Search and Data Mining}{February 27-March 3, 2023}{Singapore, Singapore}
\acmBooktitle{Proceedings of the Sixteenth ACM International Conference on Web Search and Data Mining (WSDM '23), February 27-March 3, 2023, Singapore, Singapore}
\acmPrice{15.00}
\acmDOI{10.1145/3539597.3570370}
\acmISBN{978-1-4503-9407-9/23/02}

\begin{document}

% \title{More Effective Self-training for Entity Alignment}
\title{Dependency-aware Self-training for Entity Alignment}

% \affiliation{%

\author{Bing Liu}
% \authornote{Corresponding author.}
\email{bing.liu@uq.edu.au}
\affiliation{%
  \institution{The University of Queensland}
  \streetaddress{}
  \city{}
  \state{}
  \country{}
  \postcode{}
}
\author{Tiancheng Lan}
\email{tiancheng.lan@uqconnect.edu.au}
\affiliation{%
  \institution{The University of Queensland}
  \streetaddress{}
  \city{}
  \state{}
  \country{}
  \postcode{}
}
\author{Wen Hua}
\email{w.hua@uq.edu.au}
\affiliation{%
  \institution{The University of Queensland}
  \streetaddress{}
  \city{}
  \state{}
  \country{}
  \postcode{}
}
\author{Guido Zuccon}
\email{g.zuccon@uq.edu.au}
\affiliation{%
  \institution{The University of Queensland}
  \streetaddress{}
  \city{}
  \state{}
  \country{}
  \postcode{}
}


\begin{abstract}
% broad field
Entity Alignment (EA), which aims to detect entity mappings (i.e. equivalent entity pairs) in different Knowledge Graphs (KGs), is critical for KG fusion.
% narrow the context to neural EA and self-training
Neural EA methods dominate current EA research but still suffer from their reliance on labelled mappings.
To solve this problem, a few works have explored boosting the training of EA models with self-training, which adds confidently predicted mappings into the training data iteratively.
% limitations
Though the effectiveness of self-training can be glimpsed in some specific settings, we still have very limited knowledge about it.
One reason is the existing works concentrate on devising EA models and only treat self-training as an auxiliary tool.
To fill this knowledge gap, we change the perspective to self-training to shed light on it.
% benchmark the existing self-training EA methods to support systematic studies about it.
% limitations/gaps of existing research
In addition, the existing self-training strategies have limited impact because they introduce either much False Positive noise or a low quantity of True Positive pseudo mappings.
To improve self-training for EA, we propose exploiting the dependencies between entities, a particularity of EA, to suppress the noise without hurting the recall of True Positive mappings.
Through extensive experiments, we show that the introduction of dependency makes the self-training strategy for EA reach a new level. 
The value of self-training in alleviating the reliance on annotation is actually much higher than what has been realised.
Furthermore, we suggest future study on smart data annotation to break the ceiling of EA performance. 

\end{abstract}



\begin{CCSXML}
  <ccs2012>
  <concept>
  <concept_id>10002951.10002952.10003219</concept_id>
  <concept_desc>Information systems~Information integration</concept_desc>
  <concept_significance>300</concept_significance>
  </concept>
  </ccs2012>
\end{CCSXML}
\ccsdesc[300]{Information systems~Information integration}

\keywords{Knowledge Graph, Entity Alignment, Self-training, Dependency}

\maketitle

\section{Introduction}

%%# background of KG, EA
%%* What is KG, usefulness of KG
Knowledge Graphs (KGs), which represent entities and their relationships with graphs, have been widely used as knowledge drivers for many applications, such as semantic search, entity search, and recommender systems~\cite{DBLP:journals/ftir/ReinandaMR20,DBLP:journals/ftir/ReinandaMR20,DBLP:journals/tnn/JiPCMY22}.
%%* -> Incompleteness of KG. (broad gap)
Due to the confined knowledge source and imperfect knowledge extraction techniques, a KG can hardly avoid the incompleteness issue.
To solve this problem, fusing different KGs into a more comprehensive one is thought a promising direction.
%%* -> EA
Entity Alignment (EA), which identifies equivalent entities between two KGs, is a critical step in KG fusion.

% As shown in Fig.~\ref{}, EA aims to identify the entity mappings xxxx. By doing so, we then can integrate the knowledge about xxx and xxx from two different KGs together.

\begin{figure}
    \centering
    \includegraphics[width=8.5cm]{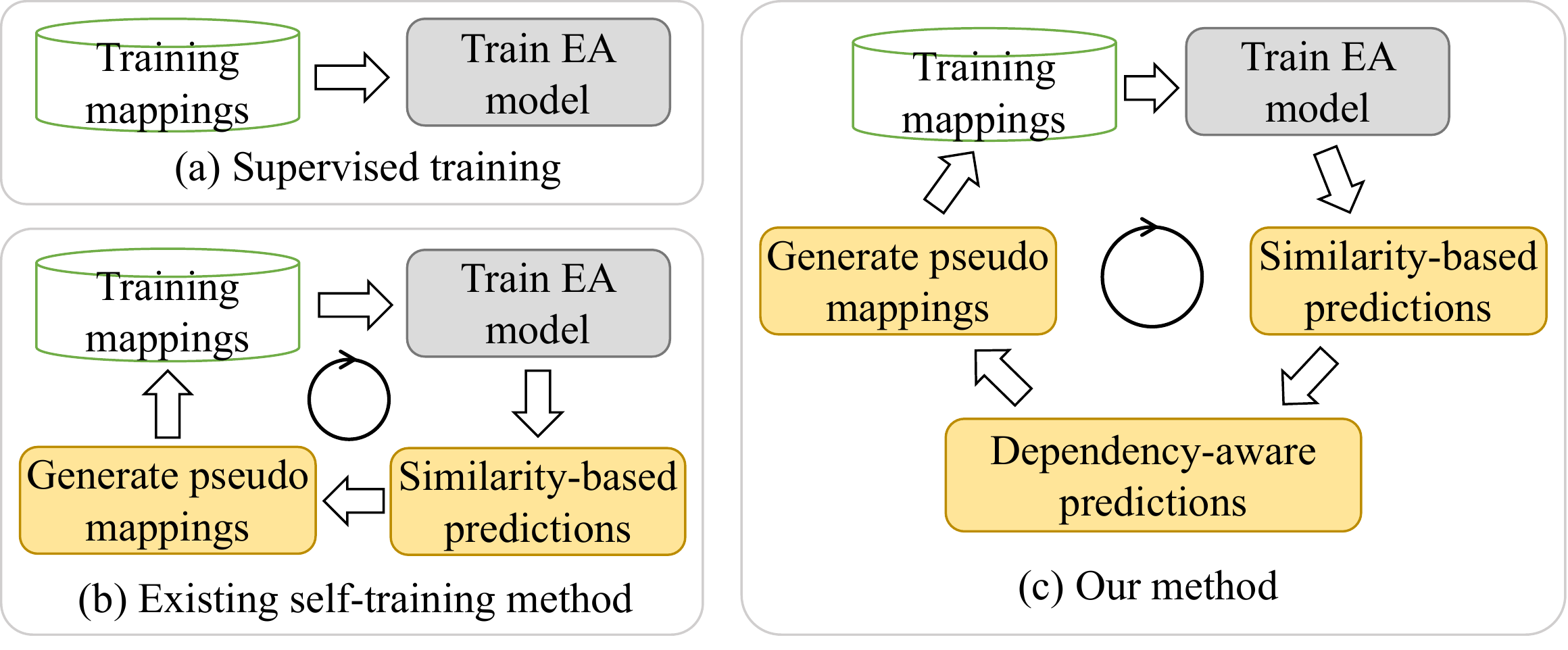}
    \vspace{-7mm}
    \caption{Illustration of self-training for EA. (a) Supervised training; (b) Self-training adds pseudo mappings into the training data iteratively; (c) Our method derives dependency-aware predictions to improve self-training.}
    \label{fig:intro}
    \vspace{-20pt}
\end{figure}

%%# Existing works of EA. -> Gap
%%* neural EA. (-> gap: training data)
%%* -> semi-supervised EA. (-> gap: quality of pseudo-mappings)
Neural models dominate current EA research.
The general idea of these methods is: encode the entities from two different KGs into the same embedding space; then, for each source entity, the target entity with the highest similarity in the embedding space is selected as its counterpart.
Effective training is critical for getting informative entity embeddings.
Supervised training relies on some pre-aligned mappings (i.e. seed mappings) as training data, which are costly to annotate (as in Fig.~\ref{fig:intro} (a)).
To improve EA effectiveness without increasing annotation cost, self-training, a category of semi-supervised learning, has been explored to train the EA models more effectively~\cite{DBLP:conf/ijcai/ZhuXLS17,DBLP:conf/ijcai/SunHZQ18,DBLP:conf/wsdm/MaoWXLW20}.
In self-training strategies, we first train the EA model with labelled data. Then, we select some predicted mappings with high confidence (i.e. pseudo mappings) to add to the training data (together with the labelled data) and update/re-train the EA model (as in Fig.~\ref{fig:intro} (b)).
Basically, the predicted mappings with high similarities are picked up~\cite{DBLP:conf/ijcai/ZhuXLS17}.
To further improve the accuracy of pseudo mappings, Sun et al.~\cite{DBLP:conf/ijcai/SunHZQ18} impose a one-to-one constraint on the derivation process;
Alternatively, Mao et al.~\cite{DBLP:conf/wsdm/MaoWXLW20} choose to use two entities (from both KGs) which are mutually nearest neighbours in the embedding space to form a pseudo mapping.
These works demonstrate that their proposed self-training strategies can significantly boost their specific EA models than supervised learning.
Though promising, the existing works about self-training for EA have apparent limitations.

Self-training is still underexplored and has never been studied systematically in EA task.
The existing works~\cite{DBLP:conf/ijcai/ZhuXLS17,DBLP:conf/wsdm/MaoWXLW20} mainly focus on designing novel EA models, while treat self-training as an auxiliary tool for improving their model performance. In other words, most attention was on the EA model but not the self-training strategies.
As a result, self-training for EA has not been well investigated, and a series of questions remain to be answered.
For example, which self-training method is the most effective? Are those self-training approaches generic across different EA models? To what extent can self-training alleviate the reliance of EA models on labelled data? and etc.
To fill this gap, we benchmark the existing self-training methods for EA  to shed light on them.

The existing self-training strategies for EA are based on the confidence of the EA model, i.e. similarities of predicted mappings. They can easily introduce \textit{False Positive} noise -- the pseudo mappings the model has high confidence in but are actually incorrect.
This kind of noise can hardly be avoided, especially when the model cannot be well-trained, such as in the few annotation settings.
As a result, the noise can propagate errors to the model in the next iteration of training and exaggerate the wrong patterns learned in previous iterations.
Unfortunately, the existing measures used to control the noise, like increasing the threshold of selecting pseudo mappings, usually come with the cost of reduced \textit{True Positive} pseudo mappings, which impairs the impact of self-training in another way.
To escape such a dilemma, we believe the heuristic for generating pseudo mappings should go beyond the confidence of EA model.

We claim that exploiting the potential dependencies between entities, which is a particularity of graph data, is promising in improving the self-training strategy for EA.
In EA task, this dependency can be interpreted as: the counterpart of one entity can affect the counterparts of its neighbouring entities.
Because neural EA models perform inference independently, it is very possible for them to make predictions that violate the dependencies between entities.
In our terminology, a set of mappings without violating any dependency are \textit{compatible}, otherwise \textit{incompatible}.
Intuitively, a set of better predictions should be more compatible.
Thus, by checking the violation of dependencies within a set of predictions, we have the chance to detect the suspicious ones and derive more compatible predictions.
This would be helpful for generating less noisy pseudo mappings in self-training, as shown in Fig.~\ref{fig:intro} (c).

In this work, we propose a \textbf{S}elf-\textbf{T}raining framework for \textbf{EA} named \textbf{\textsf{STEA}}, which can incorporate the dependencies between entities to improve the effectiveness of training EA models.
In our \methodname framework, the EA model is initially trained with labelled data.
%
% Then, given a set of similarity-based EA predictions, we define their overall compatibility as a joint distribution and model it with a graphical model, which can aggregate the local compatibilities.
% %
% Afterwards, with this compatibility model, we adjust the similarity-based predictions to be more compatible. For each entity, we derive a dependency-aware probability distribution over all counterpart candidates.
%
Then, given a set of similarity-based EA predictions, we measure their overall compatibility using a probabilistic graphical model, and adjust them to be more compatible. For each entity, we derive a dependency-aware probability distribution over all counterpart candidates. The likelihood of candidate causing low compatibility will be suppressed.
Finally, we generate pseudo mappings based on the dependency-aware EA predictions. A pair of entities having the highest probabilities on each other mutually form one pseudo mapping.
The obtained pseudo mappings are added to the training data to update the EA model in conjunction with the labelled data.
This process iterates until the EA model converges.

Our contributions can be summarised as below:
\begin{itemize}[leftmargin=*]
    \item We provide a systematic study of existing self-training methods for EA through benchmarking.
    %\item We benchmark the existing self-training methods for EA to support systematic studies of them.
    \item We propose a self-training framework which can exploit the dependencies between entities to overcome the drawbacks of model confidence-based strategies.
    \item Extensive experiments~\footnote{Our code, used data, and running scripts are released at \url{https://github.com/uqbingliu/STEA}.} have verified the effectiveness of our \methodname framework. We find dependency can be used to improve the self-training strategy for EA task significantly.
    \item We find the impact of self-training on EA models is substantially undervalued. Our \semiea framework can greatly alleviate the reliance of EA models on annotations.
\end{itemize}

%%# how we fill the gap. -> exploit compatibility to derive better predictions.
%%* different mappings should be compatible.

%%# method (high-level): how to use compatibility to derive better predictions
%%* components, procedure

%%# Summarize contributions
%%* propose to improve the pseudo-mappings with KG structure
%%* propose one framework which exploit compatibility to derive better pseudo-mappings
%%* experimental finding

\section{Related Work}

\subsection*{Neural EA Model}

Neural EA models~\cite{DBLP:journals/pvldb/SunZHWCAL20,DBLP:conf/coling/ZhangLCCLXZ20,DBLP:journals/tkde/ZhaoZTWS22,DBLP:conf/cikm/LiuHZZZ22} emerge with the development of deep learning techniques and has been the mainstream of current EA research.
KG encoders lie in the core of neural EA models.
Various neural architectures have been explored to encode entities.
The initial works tried translation-based models ~\cite{DBLP:conf/ijcai/ChenTYZ17,DBLP:conf/ijcai/ZhuXLS17}.
Later, the emergence of Graph Convolutional Network (GCN)~\cite{DBLP:journals/tnn/WuPCLZY21} changed the landscape of neural EA models.
Many GCN-based KG encoders for EA task were designed and showed significant superiority than the translation-based models~\cite{DBLP:conf/emnlp/WangLLZ18,DBLP:conf/wsdm/MaoWXLW20,DBLP:conf/aaai/SunW0CDZQ20,DBLP:conf/cikm/MaoWXWL20}.
Specifically, the GCN-based EA models can unify the way of encoding structure and attribute information naturally, while it is not easy for the translation-based models to exploit the attributes in KGs.
In addition, the GCN-based methods can achieve state-of-the-art (SOTA) performance with big leads.
In this work, we choose different GCN-based SOTA EA models to study the training effectiveness of different self-training frameworks on them.
Among previous neural EA works, some only considered structure information of KG and dedicated to the modelling of KG structure~\cite{DBLP:conf/ijcai/SunHZQ18,DBLP:conf/aaai/SunW0CDZQ20,DBLP:conf/cikm/MaoWXWL20}, while some others focus on exploiting extra information in KGs other than the structure~\cite{DBLP:conf/ijcai/WuLF0Y019,DBLP:conf/aaai/0001CRC21,DBLP:conf/emnlp/LiuCPLC20}.
The former setting is more challenging and more general since structure is the most basic information in KGs.
We follow this setting in our experiments.

\subsection*{Semi-supervised Learning for EA}

Semi-supervised learning is an approach to machine learning that combines a small amount of labelled data with a large amount of unlabelled data during training.
Self-training is a wrapper method for semi-supervised learning
\cite{DBLP:journals/corr/abs-2202-12040}. In general, the model is initially trained with labelled data. Then, it is used to assign pseudo-labels to the set of unlabelled training samples and enrich the training data and train a new model in conjunction with the labelled training set.
In EA research, different self-training strategies have been explored.
Zhu et al.~\cite{DBLP:conf/ijcai/ZhuXLS17} added all predicted mappings into training data while assigned high confidence predictions high weights in their training loss.
Sun et al.~\cite{DBLP:conf/ijcai/SunHZQ18} improved the self-training strategy further by imposing a one-to-one constraint in generating pseudo mappings.
Mao et al.~\cite{DBLP:conf/wsdm/MaoWXLW20} proposed a strategy based on the asymmetric nature of alignment directions.
If and only if two entities are mutually nearest neighbours, they form one pseudo mapping.
Compared with the above two methods, this strategy does not introduce any hyperparameter and is very simple in implementation.

Apart from self-training, co-training is another semi-supervised method explored to improve the effectiveness of EA models.
Co-training is an extension of self-training in which multiple models are trained on different sets of features and generate labelled examples for one another.
In EA task, Chen et al.~\cite{DBLP:conf/ijcai/ChenTCSZ18} performed co-training of two EA models taking different types of KG information as input.
Xin et al.~\cite{DBLP:conf/aaai/XinSHLHQ022} extended the number of EA models to $k>2$.

Our work focuses on improving self-training for EA by producing better pseudo mappings based on entity dependency in the KG.

\section{Notations \& Problem Definition}
%%* notations for KGs
Suppose we have two KGs $\mathcal{G}$ and $\mathcal{G}'$ with their corresponding entity sets $E$ and $E'$.
%%* notations for EA
Given $\mathcal{G}$, $\mathcal{G}'$ and a set of labelled entity mappings (i.e. seed mappings) $M^l = \{ (e \in E, e' \in E') \}$, EA aims to identify more potential mappings.
%%* special notations for EA used by our method

\noindent
For the convenience of explaining our method, suppose we treat $\mathcal{G}$ as the source KG while  $\mathcal{G}'$ as the target KG.
We denote the counterpart variable of each source entity $\mathrm{e} \in E$ as  $y_{e} \in E'$, while represent its assignment as $\hat{y}_e$.
For simplicity, we denote the counterpart variables of a set $E$ of entities as $y_E$ collectively, i.e. $y_E = \{ y_e | e \in E \}$.
The sets of labelled and unlabelled source entities are denoted as $L \subset E$ and $U \subset E$ respectively.
%
% It should be noticed that either KG can be treated as source KG while the other is seen as target KG.

%%* notions for neural EA model
\noindent
In a neural EA method, one model $\Theta$ can measure the similarity $\mathrm{Sim}_\Theta(e, e')$ between each source entity $e \in E$ and each target entity $e' \in E'$. The counterpart of entity $e$ can be inferred via $\hat{y}_e = \arg \max_{e' \in E'} \mathrm{Sim}_\Theta(e, e')$.
% Here, we use $\Theta$ to represent a particular EA model.

\section{The \semiea Framework}

\begin{figure}[!t]
    \centering
    \includegraphics[width=6.8cm]{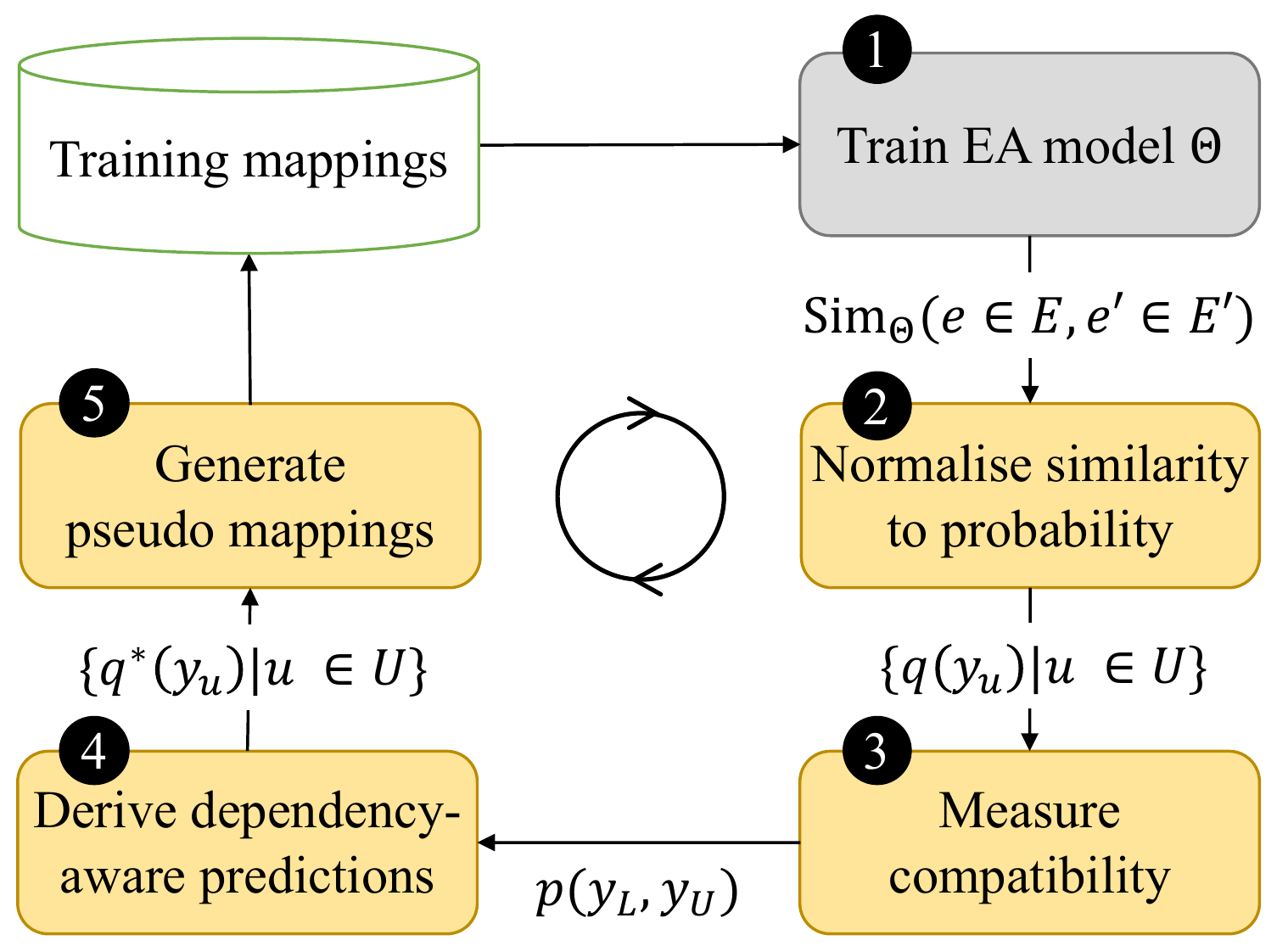}
    \caption{Overview of the \semiea framework.}
    \vspace{-8pt}
    \label{fig:em_proc}
\end{figure}

%%# components and running procedure of SemiEA
Fig.~\ref{fig:em_proc} shows an overview of our \semiea framework.
In each iteration, \semiea performs the following operations:
\circled{1} Train the neural EA model with the training mappings, which are the labelled mappings initially and will incorporate the pseudo mappings since the second iteration;
\circled{2} Normalise the similarities $\{ \mathrm{Sim}_\Theta(e, e'), \forall e' \in E' \}$ between each source entity $e$ and all target entities into probability distribution $q (y_e) = \{ \Pr(y_e = e'), \forall e' \in E' \}$.
For this purpose, a normalisation model is learned separately (Sec. \ref{sec:normalise});
\circled{3} Model the overall compatibility of current predictions $\{q(y_u), u \in U \}$ by modelling the joint probability $p(y_L, y_U)$ of all (labelled and predicted) mappings. In particular, we compute the local compatibility at each entity firstly and then aggregate all local compatibilities into a global one (Sec. \ref{sec:self-consistency});
\circled{4} Derive dependency-aware (i.e. more compatible) predictions $\{q^*(y_u), u\in U \}$ from $\{q(y_u), u\in U\}$ with the assistance of the compatibility model (Sec. \ref{sec:dependency}). 
This step is designed to suppress the likelihood of suspecious predictions.
Since either KG can be treated as the source KG while the other one is seen as the target KG, we can also derive $\{q^*(y_{u'}), u' \in U' \}$ from $\mathcal{G}'$ in the same way;
\circled{5} Generate pseudo mappings and combine them with the labelled mappings to enrich the training data. Based on the dependency-aware predictions, we explore a few different strategies for producing the pseudo mappings (Sec. \ref{sec:pseudo}).
The iteration process (i.e. \circled{1}-\circled{5}) repeats until $\Theta$ converges.

\subsection{Normalising Similarity to Probability}
\label{sec:normalise}
%%# why we need to do normalisation
Our method relies on the probabilities $q(y_e)=\{ \Pr(y_e=e'), \forall e' \in E' \}$ of counterpart candidates for each source entity $e$, while the existing neural EA models only output its similarities $\{ \mathrm{Sim}(e, e'), \forall e' \in E' \}$ with all candidates. Therefore, we need to normalise the similarities into probabilities.

%%# How
%%* why need a separate model: simple normalisation is not good
Because some simple normalisation method like MinMax scaler cannot lead to a proper probability distribution, we introduce a separate normalisation model, which is based on softmax function and has learnable parameters.
%%* method details
Specifically, for each entity $e \in E$, we first transform its corresponding similarities linearly as in Eq.~\ref{eq:simtoprob_1}, where $\omega_0, \omega_1$ are parameters.
Then, the obtained feature is fed into a softmax function with temperature factor $\tau$, as in Eq.~\ref{eq:simtoprob_2}.
In this way, we get the probabilities of all possible counterparts for entity $e$.
To learn the parameters $\Omega = \{ \omega_0,\omega_1, \tau \}$, we minimise the cross-entropy loss on the labelled data, i.e. Eq.~\ref{eq:objective_sim2prob}. Here, $\hat{y}_e | e \in L$ is the ground-truth counterpart of entity $e$.
%%* conclusion of this section. (answer so what problem)
Eventually, with the learned normalisation model, we obtain a probability distribution $q(y_u)$ over all counterpart candidates for each unlabelled entity $u \in U$.

\begin{equation}
    f (e,e') = \omega_1 \cdot \mathrm{Sim}_\Theta (e,e') + \omega_0
    \label{eq:simtoprob_1}
\end{equation}

\begin{equation}
    \mathrm{Pr}_{\Omega}(y_e=e') = \frac{\exp(f (e,e')/\tau)}{ \sum_{e'' \in E'}  \exp(f (e, e'')/\tau)  }
    \label{eq:simtoprob_2}
\end{equation}

\begin{equation}
    O_{\Omega} = - \sum_{e \in L} \log \mathrm{Pr}_{\Omega} (y_e = \hat{y}_e)
    \label{eq:objective_sim2prob}
\end{equation}

\subsection{Measuring Compatibility}\label{sec:self-consistency}
%%* local compatibilities -> global compatibility
It is not easy to directly check the compatibility of a large number of mappings $y_L, y_U$.
To solve this problem, we first measure the local counterpart compatibility at each source entity, which only involves predictions of its neighbouring entities, and then aggregate the local compatibilities using a graphical model~\cite{DBLP:journals/ftml/WainwrightJ08,bishop2006chapter}.

\subsubsection{Local Compatibility}
\begin{figure}
    \centering
    \includegraphics[width=8.4cm]{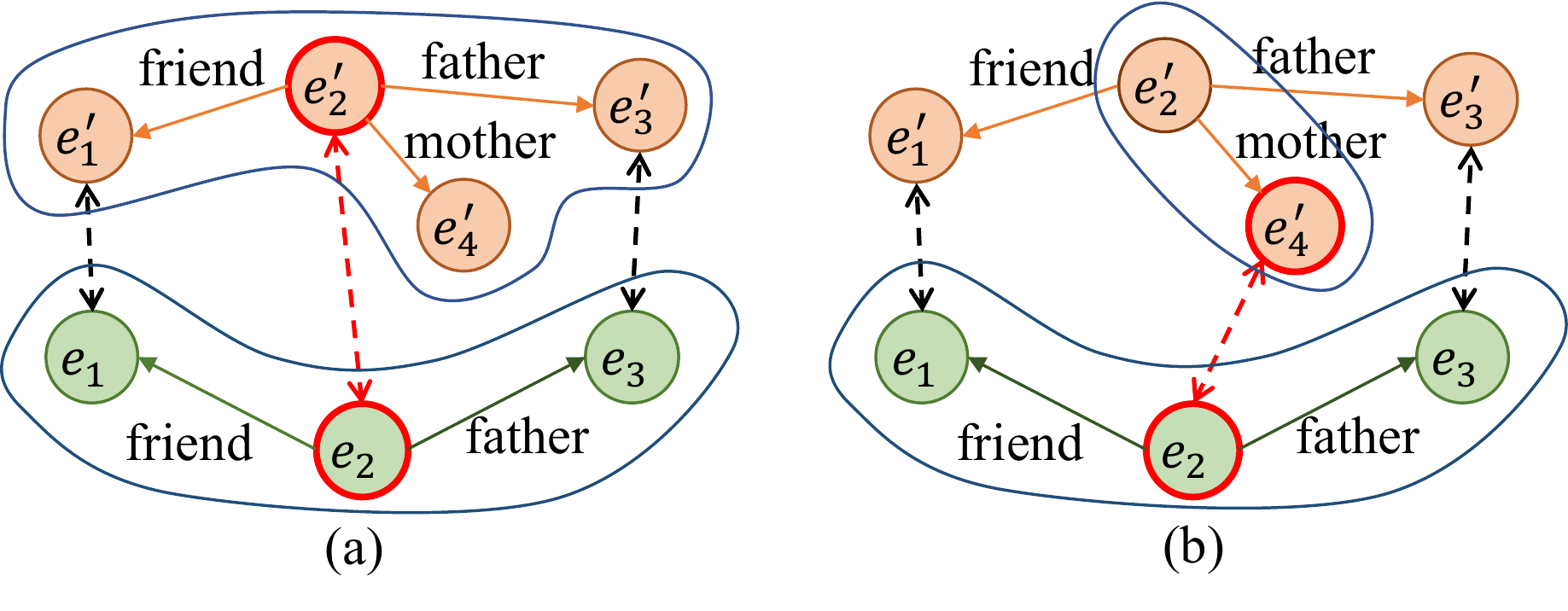}
    \vspace{-12pt}
    \caption{Example of PARIS compatibility. In (a), $y_2=e'_2$ can get supporting evidence from $y_1=e'_1$ and $y_3=e'_3$, while $y_2=e'_4$ in (b) cannot. Thus, the compatibility at $e_2$ in (a) is higher than that in (b).}
    \vspace{-10pt}
    \label{fig:paris_comp}
\end{figure}

%%* intuition of local compatibility
We check the local compatibility based on the dependencies between each entity and its neighbouring entities:
\textit{one mapping $e \equiv e'$ (i.e. $y_e=e'$) should be able to be inferred from the other mappings between their neighbours $\mathcal{N}_e$ and $\mathcal{N}_{e'}$}.
%%* one example - how to do compatibility checking based on this idea.
In Fig.~\ref{fig:paris_comp}, we use examples to show how to apply this idea for compatibility checking.
In each plot of Fig.~\ref{fig:paris_comp}, we are given two KGs to align and some predicted mappings, and suppose we need to check the local compatibility at $e_2$.
In plot (a), for the mapping $e_2 \equiv e'_2$, we can find two mappings between the neighbours of $e_2$ and $e'_2$, i.e., $e_1 \equiv e'_1$ and $e_3 \equiv e'_3$. They can provide supporting evidence for $e_2 \equiv e'_2$: if two entities have equivalent father entities and equivalent friend entities, they might also be equivalent.
However, in plot (b), the mapping $e_2 \equiv e'_4$ cannot get this kind of supporting evidence since there is no mapping between the neighbours of $e_2$ and $e'_4$.
Thus, the local compatibility at $e_2$ in plot (a) is higher than that in plot (b).

%%* formal expression of local compatibility
Based on this intuition, Suchanek et al.~\cite{DBLP:journals/pvldb/SuchanekAS11} designed a reasoning-based EA method named PARIS, which could achieve promising EA performance~\cite{DBLP:journals/tkde/ZhaoZTWS22,DBLP:journals/pvldb/SunZHWCAL20}.
In this work, we adapt the reasoning technique proposed in PARIS~\cite{DBLP:journals/pvldb/SuchanekAS11} to quantify the compatibility, as formulated in Eq.~\ref{eq:paris_prob}.
Here, $F_e$ denotes the entity set containing $e$ and its neighbours $\mathcal{N}_e$, and is called a factor subset;
$r(e,n)$ represents a certain triple with head entity $e$, relation $r$, and tail entity $n \in \mathcal{N}_e$;
$\Pr(r' \subseteq r)$ denotes the probability that $r'$ is a sub-relation of $r$, while $fun^{-1}(r)$ denotes the inverse functionality of relation $r$ (See \cite{DBLP:journals/pvldb/SuchanekAS11} for more detailed explanation).
The item $\mathbbm{1}_{y_n = n'}$ indicates whether $y_n$ equals to $n'$.
The whole equation of $g(y_e = e')$ expresses the likelihood that prediction $y_e$ can be inferred from its neighbouring predictions $y_{\mathcal{N}_e}$.

\vspace{-8pt}
\begin{equation}
    \begin{aligned}
        g(y_{F_e}) = & 1 - \prod_{\substack{r(e,n), r'(y_e,n')}}
         \left(1 - \Pr (r' \subseteq r ) \times fun^{-1}(r) \times \mathbbm{1}_{y_n = n'} \right)
         \\
        & \times  \left(1 - \Pr (r \subseteq r' ) \times fun^{-1}(r') \times \mathbbm{1}_{y_n = n'} \right)
    \end{aligned}
    \label{eq:paris_prob}
\end{equation}

%%* why I say those?
We emphasize that:
(1) When checking the local compatibility at entity $e$, only a few predictions $y_{F_e}$ are involved;
(2) The compatibility score $g(y_{F_e})$ at entity $e$ does not mean the correctness of prediction $y_e$ exactly. High compatibility is a good signal indicating $y_e$ is a correct prediction, but a low compatibility score might be caused by wrong $y_e$ or wrong neighbouring predictions $y_{\mathcal{N}_e}$.

% \begin{equation}
%     l (y_F) = \exp \left(g_\kappa(y_F)  \right)
%     \label{eq:local_func}
% \end{equation}

\subsubsection{Overall Compatibility}\label{sec:overall_compatibility}

We further formulate the overall compatibility by aggregating local compatibilities at all entities $e \in E$ as in Eq.~\ref{eq:overall_comp}, where $z$ is for normalisation.

\begin{equation}
    p (y_L, y_U) = \frac{1}{z} \prod_{e \in E} \exp \left( {g(y_{F_e})} \right), z = \sum_{\hat{y}_U \in {E'}^{|U|}} \prod_{e \in E} \exp \left( {g(\hat{y}_{F_e})} \right)
    % = \frac{1}{z} \exp \left( \sum_{F \in \mathcal{F}} \left(\sum_{r \in R} \alpha_r \cdot g_r(y_F) + \alpha_0 \right) \right)
    \label{eq:overall_comp}
\end{equation}

% \begin{equation}
%     z = \sum_{y_U \in E^{|U|}} \prod_{e \in E} \mathrm{e}^{g(y_e)}
% \end{equation}

\noindent
Note that $z$ is intractable because it involves an integral over $y_U \in {E'}^{|U|}$, which is a very large space. For such computation reason, we avoid computing $p_\Phi (y_L, y_U)$, conditional probability like $p_\Phi (y_U|y_L)$, and marginal probability $p_\Phi (y_L)$ directly in the following sections.

% Instead, we will exploit $p(y_e|y_{-e})$ ($-e$ refers to $E \setminus e$), whose computation is actually much easier.
% As in Eq.~\ref{eq:cond_p}, computing $p(y_e|y_{-e})$ only involves a few factor subsets containing $e$.

% $\mathrm{MB}^e$ is the Markov Blanket of $e$, which only contains entities cooccuring in any factor subset with $e$.

% Because a certain entity $e$ is only involved in few factor subsets $\mathcal{F}^e = \{F \in \mathcal{F} | e \in F \}$, $p_\Phi(y_e|y_{-e})$ can be simplified as in Eq.~\ref{eq:cond_p}, where $\mathrm{MB}^e$ is the Markov Blanket of entity $e$ which is formed by entities co-occuring with $e$ in any $F \in \mathcal{F}^e$
% %
% Thus, only entities in $\mathrm{MB}^e$ need to participate in computing $p_\Phi(y_e | y_{\mathrm{MB}^e} )$.

% \begin{equation}
%     \begin{aligned}
%         p_\Phi(y_e | y_{-e})  = p_\Phi(y_e | y_{\mathrm{MB}^e}) \\
%         = \frac{\prod_{F | e \in F} l(y_F) }{\sum_{e' \in E'} \prod_{F | e \in F} l(y_F | y_e=e') }
%     \end{aligned}
%     \label{eq:cond_p}
% \end{equation}

% \subsection{Improving Predictions w.r.t. Compatibility}

\subsection{Deriving Dependency-aware Predictions}
\label{sec:dependency}

% The neural EA model infers the counterpart of each source entity independently while neglecting the dependencies between different entities.

% Intuitively, good predictions should be compatible.

% We adjust the predictions so that they can be more compatible.

% Our goal is to get a set of mappings $y_U$ leading to high overall compatibility.
% If we have $p(y_U | y_L)$, sampling $y_U$ from it would be a straightforward solution.
% %%* how to:
% However, it is intractable to derive the $p(y_L, y_U)$ and $p(y_U | y_L)$, as mentioned in Sec.~\ref{sec:overall_compatibility}.
% %
% To solve this problem, we try to derive a distribution $q^*(y_U)$ which can approximate the posterior distribution $p(y_U | y_L)$. Once we get such a $q^*(y_U)$, we can use it inplace of $p(y_U | y_L)$.

% To this end, we first choose to use the distribution $q(y_U)$ predicted by EA model to approximate $p(y_U | y_L)$.
% Because the neural EA model makes inference independently, we can factorise it as in $q(y_U) = \prod_{u \in U} q(y_u)$.
%

Precise predictions should have high overall compatibility.
% To improve the compatibility of predictions $q(y_U)$
To this end, we try to update $q(y_U)$ so as to minimise the KL-divergence $\mathrm{KL}(q(y_U) || p(y_U|y_L))$.
However, it is difficult to minimise the KL-divergence directly since we even cannot derive $p(y_U|y_L)$ as mentioned in Sec.~\ref{sec:overall_compatibility}.
To overcome this problem, we apply variational inference~\cite{DBLP:journals/ijon/Hogan02}, in which we optimise the Evidence Lower Bound (ELBO) of the KL-divergence.
As shown in Eq.~\ref{eq:KL} and \ref{eq:elbo}, the KL-divergence can be written as the difference between observed evidence $\log p (y_L)$ and its ELBO.
Thus, minimising the KL-divergence is equivalent to maximising the ELBO, which is computationally simpler as will be seen below.

\vspace{-8pt}
\begin{equation}
    \begin{aligned}
        \mathrm{KL}(q(y_U) || p(y_U|y_L)) =  \log p(y_L)
        - \mathrm{ELBO}
    \end{aligned}
    \label{eq:KL}
\end{equation}

\vspace{-8pt}
\begin{equation}
    \mathrm{ELBO} = \mathbb{E}_{q (y_U)} \log p(y_U, y_L) - \mathbb{E}_{q (y_U)} \log q (y_U)
    \label{eq:elbo}
\end{equation}

Towards deriving an easy-to-use form of ELBO, we first notice $q(y_U)$ can be factorised as in $q(y_U) = \prod_{u \in U} q(y_u) $ because the neural EA model infers $y_u$ for $u \in U$ independently.
In addition, we use pseudolikelihood~\cite{besag1975statistical} to approximate the joint probability $p(y_U, y_L)$ for simpler computation, as in Eq.~\ref{eq:pseudolikelihood}.
Here, the computation of conditional distribution $p(y_u | y_{E \setminus u})$ can be simplified as in Eq.~\ref{eq:cond_p}~\footnote{The derivation process can be done by introducing Eq.~\ref{eq:overall_comp} and reducing the items $g(F_i)$ irrelevant to $y_u$.}, where $\mathrm{MB}_u = \bigcup_{u \in F_i} F_i  $ is called the Markov Blanket of $u$ and actually contains $u$ and its two-hop neighbours.

% \vspace{-8pt}
\begin{equation}
    \begin{aligned}
        p(y_U, y_L) & = p(y_L) \cdot \prod_{i = 1... |U|} p(y_{U_i} |  y_{U_{1:i-1}, y_L} ) \\
        & \approx p(y_L) \cdot \prod_{u \in U} p(y_u | y_{E \setminus u} )
    \end{aligned}
\label{eq:pseudolikelihood}
\end{equation}

\begin{equation}
    \begin{aligned}
        &p(y_u | y_{E \setminus u}) = \frac{p(y_E)}{\sum_{e' \in E'} p(y_u=e', y_{E \setminus u})}  \\
        & = \frac{ \prod_{F_i | u \in F_i} \exp \left( g(y_{F_i}) \right) }{\sum_{e' \in E'} \prod_{F_i | u \in F_i} \exp \left(g(y_{F_i}|y_u=e') \right) } \doteq p(y_{u} | y_{\mathrm{MB}_u})
        % & \doteq p(y_e | y_{\mathrm{MB}^e})
    \end{aligned}
    \label{eq:cond_p}
\end{equation}

\noindent
Then, we can derive an approximation of ELBO shown in Eq.~\ref{eq:elbo_approx} by introducing the factorised $q(y_U)$, Eq.~\ref{eq:pseudolikelihood}, and Eq.~\ref{eq:cond_p} to Eq.~\ref{eq:elbo}.
% \todo{derivation process is not clear}
% (see Appendix~\ref{app:Q_proc} for derivation details),

% \vspace{-8pt}
\begin{equation}
    \begin{split}
        Q = \log p(y_L) + \sum_{u \in U} \mathbb{E}_{q(y_u)}
         \Big[ \mathbb{E}_{q({y_{\mathrm{MB}_u }})}
         [ \log p(y_u | y_{\mathrm{MB}_u }) ]
         - \log q(y_u)   \Big]
    \end{split}
    \label{eq:elbo_approx}
\end{equation}

Now, our goal becomes to get optimal $q(y_u)$  which maximises $Q$.
We solve this problem with coordinate ascent~\cite{wright2015coordinate}.
In particular, we update $q(y_u)$ for each $u \in U$ in turn iteratively. Everytime we only update a single (or a block of) $q(y_u)$ with Eq.~\ref{eq:q_star}, which can be derived from $\frac{d Q}{d q(y_u)} = 0$, while keeping the other $q(y_{e \in U \setminus u})$ fixed.
This process ends until $q(y_{e \in U})$ converges.

% \vspace{-8pt}
\begin{equation}
    q^*(y_u) \propto \exp \left(\mathbb{E}_{q(y_{ \mathrm{MB}_u })}  \log p(y_u | y_{\mathrm{MB}_u})  \right)
    \label{eq:q_star}
\end{equation}

% With the obtained $q^*(y_u)$, we can sample predicted mappings  for $u \in U$ easily via $\hat{y}_u = \arg \max_{e' \in E'} q^*(y_u=e')$.

Eventually, the original $\{q(y_u) | u \in U\}$ are adjusted to $\{q^*(y_u) | u \in U \}$ for better compatibility.

\vspace{-4pt}
\subsection{Generating Pseudo Mappings}
\label{sec:pseudo}

As mentioned, we can derive $\{q^*(y_u), u \in U \}$ and $\{q^*(y_{u'}), u' \in U \}$ from both KGs.
Based on them, we explore three types of strategies for generating pseudo mappings.

% \vspace{-4pt}
\subsubsection{UniThr}
In this strategy, we generate the pseudo mappings in unidirection based on probability threshold.
Suppose choose $\mathcal{G}$ as the source KG while $\mathcal{G}'$ as the  target KG.
% We can derive  $q^*(y_U)$.
%
For each $u \in U$, we sample $\hat{y}_u = \arg \max_{e' \in E'} q^*(y_u=e')$. If $q^*(\hat{y}_u)$ is greater than a threshold $\alpha$, $(u, \hat{y}_u)$ forms a pseudo mapping.
When using this strategy, we would need to search two hyperparameters -- the choice of source KG and a probability threshold.

% \vspace{-20pt}
\subsubsection{BiThr}
In this strategy, we generate the pseudo mappings in bidirection using a probability threshold.
After deriving the two groups of dependency-aware predictions, we use one threshold to filter some predictions as pseudo mappings for each group as in \textit{UniThr}.
The obtained two sets of pseudo mappings are merged directly.
Compared with \textit{UniThr}, \textit{BiThr} only has a threshold as its hyperparameter.

\subsubsection{MutHighestProb}
As in \textit{BiThr}, we exploit both  $\{q^*(y_u), u\in U\}$ and $\{q^*(y_{u'}, u' \in U')\}$.
If two entities $u$ and $u'$ mutually have the highest probability on each other, i.e. $u' = \arg \max q^*(y_u)$ and $u = \arg \max q^*(y_{u'})$, then we select $e \equiv e'$ as one pseudo mapping.
This strategy has no hyperparameter and thus would be easy to use in practice.

In our framework, the strategy \textit{MutHighestProb} is used by default, while the others are explored for comparison.

\subsection{Implementation}

\begin{algorithm}[t!]
    \caption{The \semiea Framework}
    \label{alg}
    Train neural EA model $\Theta$ using labelled data $M^l$ \;
    \For{iterations}{
        Normalise EA similarity to get $q (y_u)$ \tcp{Eq.~\ref{eq:simtoprob_1}, \ref{eq:simtoprob_2}, \ref{eq:objective_sim2prob}}
        \tcp{Use $\mathcal{G}$ as the source KG to derive dependency-aware EA predictions}
        % Sample $\hat{y}_u \sim q(y_u)$ for $u \in U$ \;
        Measure local compatibilities at each $e \in E$ ; \tcp{Eq.~\ref{eq:paris_prob}}
        Compute $p(y_u | y_{\mathrm{MB}_u})$ for each $u \in U$; \tcp{Eq.~\ref{eq:cond_p}}
        Derive $q^*(y_u)$ for each $u \in U$; \tcp{Eq.~\ref{eq:q_star}}
        \tcp{Use $\mathcal{G}'$ as the source KG to derive dependency-aware EA predictions similarly}
        Generate pseudo-mappings $M^p$; \tcp{MutHighestProb}
        Update EA model $\Theta$ with training set $M^l \cup M^p$ \;
    }
\end{algorithm}

% In our implementation,
We take a few methods to simplify the computation.
(1) In Eq.~\ref{eq:q_star}, it is costly to estimate distribution $q^*(y_u)$ because $y_u$'s  assignment space $E'$ can be very large.
% To save computation cost,
Instead, we only estimate $q^*(y_u)$ for the top $K$ most likely candidates according to current $q(y_u)$.
% estimate $q^*(y_i)$ for the top $K$ candidates according to $q_\Theta(y_i)$, while taking probability of other assignments from $q_\Theta$.
(2) Both Eq.~\ref{eq:elbo_approx} and Eq.~\ref{eq:q_star} involve sampling from $q(y_u)$ for estimating the expectation. We only sample one $y_u$ as in $\hat{y}_u = \arg \max_{e' \in E} q(y_u=e')$ for each $u \in U$.
(3) When computing $q^*(y_U)$ with coordinate ascent, we treat $U$ as a single block and update $q^*(y_U)$ for once.

The whole process of \semiea and associated equations of each step are described in Alg.~\ref{alg}.

\vspace{-4pt}
\section{Experimental Settings}

\begin{table}
    \centering
    \caption{Performance of neural EA models. Only the structural information of KGs is used. 30\% labelled data.}
    \vspace{-8pt}
    \scalebox{0.92}[0.92]{
    \begin{tabular}{|c|c|c|c|c|c|c|}
    \hline
    \multirow{2}{*}{Method} & \multicolumn{2}{c|}{zh\_en} & \multicolumn{2}{c|}{ja\_en} & \multicolumn{2}{c|}{fr\_en} \\
    % \hline
     & Hit@1 & MRR & Hit@1 & MRR & Hit@1 & MRR \\
    \hline
    IPTransE~\cite{DBLP:conf/ijcai/ZhuXLS17} & 0.406 & 0.516 & 0.367 & 0.474 & 0.333 & 0.451 \\
    GCN-Align~\cite{DBLP:conf/emnlp/WangLLZ18} & 0.413 & 0.549 & 0.399 & 0.546 & 0.373 & 0.532 \\
    MuGNN~\cite{DBLP:conf/acl/CaoLLLLC19} & 0.494 & 0.611 & 0.501 & 0.621 & 0.495 & 0.621 \\
    RSN~\cite{DBLP:conf/icml/GuoSH19} & 0.508 & 0.591 & 0.507 & 0.590 & 0.516 & 0.605 \\
    AliNet~\cite{DBLP:conf/aaai/SunW0CDZQ20} & 0.539 & 0.628 & 0.549 & 0.645 & 0.552 & 0.657 \\
    MRAEA~\cite{DBLP:conf/wsdm/MaoWXLW20} & 0.638 & 0.736 & 0.646 & 0.735 & 0.666 & 0.765 \\
    PSR~\cite{DBLP:conf/cikm/MaoWWL21} &  0.702 & 0.781 & 0.698 & 0.782 & 0.731 & 0.807 \\
    RREA~\cite{DBLP:conf/cikm/MaoWXWL20} &  0.715 & 0.794 & 0.713 & 0.793 & 0.739 & 0.816 \\
    Dual-AMN~\cite{DBLP:conf/www/MaoWWL21} &  0.731 & 0.799 & 0.726 & 0.799 & 0.756 & 0.827 \\
    % \hline
    % BootEA (semi)~\cite{DBLP:conf/ijcai/SunHZQ18} & 0.629 & 0.703 & 0.622 & 0.701 & 0.653 & 0.731 \\
    % MRAEA (semi)~\cite{DBLP:conf/wsdm/MaoWXLW20}& 0.757 & 0.827 & 0.758 & 0.826 & 0.781 & 0.849 \\
    % PSR (semi)~\cite{DBLP:conf/cikm/MaoWWL21} &  0.802 & 0.851 & 0.803 & 0.852 & 0.828 & 0.874 \\
    % RREA (semi)~\cite{DBLP:conf/cikm/MaoWXWL20} & 0.801 & 0.857 & 0.802 & 0.858 & 0.827 & 0.881 \\
    % Dual-AMN (semi)~\cite{DBLP:conf/www/MaoWWL21} &  0.808 & 0.857 & 0.801 & 0.855 & 0.840 & 0.888 \\
    \hline
\end{tabular}
    }
    \label{tab:baselines}
    \vspace{-10pt}
\end{table}

%We set up the experiments to verify our proposed EMEA framework can effectively improve the neural EA models with reasoning rules.

\subsection{Datasets and Partitions}

We choose five datasets widely used in previous EA research. Each dataset contains two KGs and a set of pre-aligned entity mappings.
Three datasets are from \textit{DBP15K}~\cite{DBLP:conf/semweb/SunHL17}, which are cross-lingual and built from different language versions of DBpedia: French-English (\textit{fr\_en}), Chinese-English (\textit{zh\_en}), and Japanese-English (\textit{ja\_en}). 
Within each dataset, each KG contains around 20K entities, among which 15K are pre-aligned.
The other two datasets are from \textit{DWY100K}~\cite{DBLP:conf/ijcai/SunHZQ18}, each of which contains two mono-lingual KGs extracted from different sources: \textit{dbp\_yg} extracted from DBpedia and Yago, and \textit{dbp\_wd} extracted from DBpedia and Wikidata. 
Within each dataset, each KG contains 100K entities which are all pre-aligned.
Our experiment settings only consider the structural information of KGs and thus will not be affected by the problems of attributes like name bias in these datasets~\cite{DBLP:conf/emnlp/LiuCPLC20,DBLP:journals/tkde/ZhaoZTWS22}.

Most existing EA works use 30\% of the pre-aligned mappings as training data, which however was pointed out unrealistic in practice~\cite{DBLP:conf/coling/ZhangLCCLXZ20}. 
In this work, to thoroughly evaluate the self-training methods, we create a few variants of each dataset with different amounts of labelled data -- 1\%, 5\%, 10\%, 20\%, and 30\% of pre-aligned mappings, which are sampled randomly.
Within each dataset, all the remaining mappings except the labelled data form the test dataset.

% \begin{table*}
%     \centering
%     \footnotesize
%     \caption{Overall performance of \textsf{EMEA} and its comparable methods in combining supervised RREA and PARIS rule. Bold indicates best for the specific annotation percentage; all differences between RREA and other baselines are statistically significant ($p<0.01$); the hyphen '-' means not applicable because the corresponding methods do not formulate EA as a ranking problem.}
%     \vspace{-8pt}
%     \scalebox{0.79}[0.79]{
%     \input{sections/tables/overall.tex}
%     }
%     \label{tab:overall_perf}
%     \vspace{-8pt}
% \end{table*}

% \vspace{-10pt}
\subsection{Metrics}

The EA methods typically output a ranked list of candidate counterparts for each entity. Therefore, we choose metrics for measuring the quality of ranking.
Following most existing works, we use \textit{Hit@k (k=1,10)} and \textit{Mean Reciprocal Rank (MRR)} as metrics. 
\textit{Hit@k} is the proportion of entities whose ground-truth counterparts rank in top $k$ positions.
MRR assigns each prediction result the score $1/Rank(ground\mbox{-}truth\mbox{\hspace{2pt}}counterpart)$ and then averages them.
% MR is the average position of ground-truth counterparts in the prediction result.
% Hit@k and MRR mainly summarise the ranking performance for shallow positions, while MR is a deeper metric.
% Hit@1, MRR, and MR reflect the model performance at suggesting a single entity, a handful of entities, and many entities.
Higher \textit{Hit@k} or MRR indicates better performance. Statistical significance is performed using paired two-tailed t-test.

\begin{table*}[!ht]
    \centering
    \footnotesize
    \vspace{-10pt}
    \caption{Overall performance of \semiea and the baselines when running with Dual-AMN. Self-training methods can always achieve much better effectiveness than supervised training; Our method \methodname outperforms the baselines significantly; All differences between \methodname and other baselines are statistically significant (p < 0.01) except a few cells marked with $*$;}
    \vspace{-8pt}
    \scalebox{1}[1]{

\begin{tabular}{|c|c|c|c|c|c|c|c|c|c|c|c|c|c|c|c|c|}
    \hline
    \multirow{2}{*}{Anno.} & \multirow{2}{*}{Method} &\multicolumn{3}{c|}{zh\_en}&\multicolumn{3}{c|}{fr\_en}&\multicolumn{3}{c|}{ja\_en}&\multicolumn{3}{c|}{dbp\_wd}&\multicolumn{3}{c|}{dbp\_yg}\\
     & & Hit@1&Hit@10&MRR&Hit@1&Hit@10&MRR&Hit@1&Hit@10&MRR&Hit@1&Hit@10&MRR&Hit@1&Hit@10&MRR\\
     \hline
     \multirow{5}{*}{1\%}
     &Supervised&0.139&0.402&0.226&0.121&0.394&0.211&0.097&0.311&0.170&0.329&0.640&0.433&0.151&0.377&0.227\\
     \cdashline{2-17}
     &SimThr&0.299&0.594&0.403&0.215&0.535&0.325&0.166&0.414&0.250&0.684&0.887&0.755&0.369&0.658&0.477\\
     &OneToOne&0.372&0.656&0.471&0.296&0.611&0.403&0.213&0.486&0.304&0.668&0.890&0.751&0.366&0.700&0.487\\
     &MutNearest&0.338&0.644&0.446&0.273&0.604&0.387&0.204&0.477&0.295&0.670&0.901&0.760&0.351&0.658&0.462\\
     \cdashline{2-17}
     &\methodname&\textbf{0.723}&\textbf{0.861}&\textbf{0.770}&\textbf{0.584}&\textbf{0.803}&\textbf{0.657}&\textbf{0.471}&\textbf{0.700}&\textbf{0.547}&\textbf{0.866}&\textbf{0.955}&\textbf{0.898}&\textbf{0.718}&\textbf{0.847}&\textbf{0.761}\\
     \hline
     \multirow{5}{*}{5\%}
     &Supervised&0.381&0.708&0.491&0.369&0.750&0.495&0.312&0.648&0.423&0.726&0.920&0.795&0.517&0.799&0.614\\
     \cdashline{2-17}
     &SimThr&0.528&0.807&0.629&0.519&0.839&0.636&0.458&0.763&0.564&0.835&0.957&0.879&0.591&0.846&0.686\\
     &OneToOne&0.607&0.844&0.691&0.604&0.871&0.698&0.543&0.822&0.641&0.818&0.953&0.867&0.709&0.909&0.781\\
     &MutNearest&0.576&0.832&0.668&0.590&0.876&0.695&0.511&0.806&0.615&0.835&0.960&0.882&0.686&0.892&0.761\\
     \cdashline{2-17}
     &\methodname&\textbf{0.798}&\textbf{0.910}&\textbf{0.837}&\textbf{0.799}&\textbf{0.933}&\textbf{0.846}&\textbf{0.718}&\textbf{0.877}&\textbf{0.771}&\textbf{0.882}&\textbf{0.966}&\textbf{0.912}&\textbf{0.823}&\textbf{0.928}&\textbf{0.859}\\
     \hline
     \multirow{5}{*}{10\%}
     &Supervised&0.547&0.833&0.646&0.542&0.861&0.652&0.497&0.813&0.605&0.790&0.947&0.846&0.626&0.871&0.712\\
     \cdashline{2-17}
     &SimThr&0.631&0.872&0.719&0.661&0.908&0.754&0.597&0.863&0.694&0.856&0.966&0.896&0.641&0.879&0.729\\
     &OneToOne&0.691&0.889&0.760&0.717&0.922&0.790&0.664&0.893&0.744&0.839&0.961&0.884&0.744&0.926&0.810\\
     &MutNearest&0.678&0.886&0.753&0.715&0.926&0.795&0.654&0.886&0.738&0.860&0.967&0.899&0.739&0.912&0.802\\
     \cdashline{2-17}
     &\methodname&\textbf{0.806}&\textbf{0.913}&\textbf{0.844}&\textbf{0.840}&\textbf{0.951}&\textbf{0.880}&\textbf{0.778}&\textbf{0.918}&\textbf{0.826}&\textbf{0.890}&\textbf{0.971}&\textbf{0.919}&\textbf{0.848}&\textbf{0.944}&\textbf{0.881}\\
     \hline
     \multirow{5}{*}{20\%}
     &Supervised&0.663&0.899&0.747&0.686&0.921&0.771&0.649&0.903&0.739&0.831&0.963&0.879&0.717&0.921&0.791\\
     \cdashline{2-17}
     &SimThr&0.710&0.914&0.786&0.739&0.940&0.815&0.698&0.919&0.780&0.873&$0.975^*$&0.910&0.721&0.918&0.795\\
     &OneToOne&0.759&0.923&0.818&0.798&0.948&0.853&0.757&0.927&0.818&0.857&0.968&0.898&0.756&0.930&0.819\\
     &MutNearest&0.772&0.924&0.828&0.799&0.954&0.857&0.743&0.924&0.810&0.880&$0.973^*$&0.914&0.802&0.942&0.853\\
     \cdashline{2-17}
     &\methodname&\textbf{0.839}&\textbf{0.936}&\textbf{0.874}&\textbf{0.865}&\textbf{0.960}&\textbf{0.900}&\textbf{0.816}&\textbf{0.939}&\textbf{0.859}&\textbf{0.893}&\textbf{0.975}&\textbf{0.922}&\textbf{0.879}&\textbf{0.960}&\textbf{0.908}\\
     \hline
     \multirow{5}{*}{30\%}
     &Supervised&0.725&0.921&0.797&0.750&0.944&0.822&0.722&0.932&0.798&0.853&0.970&0.896&0.765&0.942&0.830\\
     \cdashline{2-17}
     &SimThr&0.750&0.927&0.817&0.779&0.951&0.845&0.740&0.930&0.811&0.883&$0.978^*$&0.918&0.769&0.937&0.833\\
     &OneToOne&0.789&0.936&0.842&0.830&0.958&0.877&0.797&0.948&0.851&0.874&0.973&0.910&0.797&0.947&0.853\\
     &MutNearest&0.803&0.936&0.852&0.831&0.961&0.880&0.793&0.942&0.848&0.893&$0.978^*$&0.924&0.841&0.957&0.885\\
     \cdashline{2-17}
     &\methodname&\textbf{0.854}&\textbf{0.942}&\textbf{0.885}&\textbf{0.875}&\textbf{0.968}&\textbf{0.909}&\textbf{0.843}&\textbf{0.950}&\textbf{0.882}&\textbf{0.906}&\textbf{0.978}&\textbf{0.932}&\textbf{0.893}&\textbf{0.965}&\textbf{0.919}\\     
    \hline
\end{tabular}

    }
    \label{tab:overall_perf}
    \vspace{-8pt}
\end{table*}

% \vspace{-10pt}
\subsection{Baselines}
We select the following three self-training methods as baselines:

\noindent
\textbf{SimThr.}
It is a general strategy to select the most confident predictions as pseudo-labelled data.
Zhu et al.~\cite{DBLP:conf/ijcai/ZhuXLS17} followed this idea and implemented it with weighted loss, which makes the training method coupled with the EA model.
Instead, we implement it in a typical way -- select the predicted mappings with similarities over a certain threshold and add them to the training data. Obviously, no EA-specific measure is taken to improve the quality of pseudo mappings in this strategy.

\noindent
\textbf{OneToOne.}
The work BootEA~\cite{DBLP:conf/ijcai/SunHZQ18} proposed to improve the quality of pseudo-mappings by introducing a one-to-one constraint.
In each iteration, BootEA filtered the candidate mappings with a similarity threshold and derived mappings with Maximum Bipartite Matching under the one-to-one constraint.
The generated pseudo mappings in different iterations are accumulated, while conflicting mappings violating the one-to-one constraint are resolved.
% Compared with \textit{SimThr}, this strategy is more complex.

\noindent
\textbf{MutNearest.}
In the work MRAEA~\cite{DBLP:conf/wsdm/MaoWXLW20}, Mao et al. applied a simple strategy:
if and only if the entities $e \in E$ and $e' \in E'$ are mutually nearest neighbours of each other, then the pair $(e,e')$ is considered as a pseudo mapping. 
Compared with \textit{SimThr} and \textit{OneToOne}, this strategy does not introduce any hyperparameter.

\vspace{-5pt}
\subsection{Neural EA models}
We apply each self-training method to different EA models to evaluate its effectiveness and generality.
As for our choice of EA models, we select Dual-AMN~\cite{DBLP:conf/www/MaoWWL21}, RREA~\cite{DBLP:conf/cikm/MaoWXWL20}, AliNet~\cite{DBLP:conf/aaai/SunW0CDZQ20}, and GCN-Align~\cite{DBLP:conf/emnlp/WangLLZ18}.
These neural models vary in performance, KG encoders, and etc. Among them, Dual-AMN and RREA are SOTA models.
See Table~\ref{tab:baselines} for a performance summary of existing EA models.

\vspace{-5pt}
\subsection{Details for Reproducibility}

\noindent
\textbf{Hyperparameters}.
We search the number of candidate counterparts $K$ from [5,10,15,20,25], and set it as 10 for the trade-off of effectiveness and efficiency.

\noindent
\textbf{Implementation of Baselines and Neural EA Models}.
The baseline \textit{OneToOne} is implemented by referring to the source code of BootEA implemented in OpenEA~\footnote{\url{https://github.com/nju-websoft/OpenEA}}, while the other baselines are implemented by ourselves. The implementation of all baselines is also included in our released source code.
For the neural EA models, Dual-AMN~\footnote{\url{https://github.com/MaoXinn/Dual-AMN}}, RREA~\footnote{\url{https://github.com/MaoXinn/RREA}}, and GCN-Align~\footnote{https://github.com/1049451037/GCN-Align} are implemented based on their source codes, while AliNet is implemented with OpenEA.
We use the default settings of their hyperparameters in these source codes. 
For a fair comparison, all the self-training methods run for the same number of iterations and the same number of epochs within each iteration under a certain experimental setting.

\noindent
\textbf{Configuration of Running Device}.
The experiments on 15K datasets were run on one GPU server, which is configured with an Intel(R) Xeon(R) Gold 6128 3.40GHz CPU, 128GB memory, 3 NVIDIA GeForce GTX 2080Ti GPUs and Ubuntu 20.04 OS.
The experiments on 100K datasets were run on one computing cluster, which runs CentOS 7.8.2003, and allocates us 200GB memory and 2 NVidia Volta V100 SXM2 GPUs.

\section{Results and Analysis}

\begin{figure*}[!ht]
    %\begin{minipage}{1\textwidth}
        \centering
        \vspace{-10pt}
        \includegraphics[width=17.9cm]{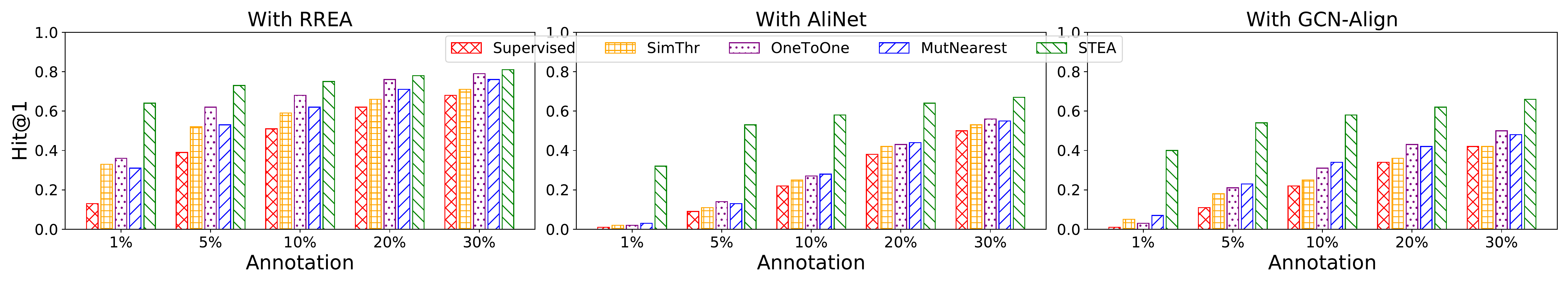}
        \vspace{-20pt}
        %\text{(a) On dataset zh\_en.}
    %\end{minipage}
    %\begin{minipage}{1\textwidth}
        %\centering
        %\includegraphics[width=17.9cm]{images/generality_ea_model_bar_jaen.pdf}
        %\text{(b) On dataset ja\_en.}
    %\end{minipage}
    \caption{The performance of self-training methods running with another three EA models: RREA, AliNet, and GCN-Align. The superiority of \methodname is genetic across different EA models.}
    \label{fig:generality_neural_models}
    \vspace{-10pt}
\end{figure*}

%%# Can SemiEA achieve better effectiveness?
%%* overall performance. a big table. EA model? different percents;  
%%* generality across neural models. (Dual-AMN, RREA, AliNet). curve figures.
%%# Examine SemiEA: Precision, Recall, F1 of pseudo-mappings
%%# Effect of Parameters. top K.

\subsection{Comparing \methodname with the Baselines}

\subsubsection*{Overall performance}
% compare with baselines
% compare different variants of STEA
To compare \methodname with the baselines, we run these different self-training methods with the SOTA EA model Dual-AMN~\cite{DBLP:conf/www/MaoWWL21} on five datasets using different annotation amounts ranging from 1\% to 30\%.
Table~\ref{tab:overall_perf} reports their results measured with three metrics. Our observations include:
(1) By comparing the supervised training with each self-training method, we can see that all the self-training methods can bring very significant improvement to the final EA effectiveness under all the settings. Thus, self-training is a better way of training EA model. 
(2) OneToOne shows an obvious advantage over SimThr, which reveals that one-to-one constraint is useful for deriving pseudo mappings of higher quality.
(3) MutNearest outperforms SimThr consistently, while both MutNearest and OneToOne have their own merits in different experimental settings. Note that MutNearest has the advantage of being easier to use in practice.
(4) Comparing \methodname with each baseline, we can observe that  \methodname has an obvious advantage in performance under all the settings. The success of \methodname verifies it is promising to exploit the dependencies between entities, which is a characteristic of graph data, in devising the self-training framework for EA.

\noindent
To conclude, self-training can always provide much more effective training for the SOTA EA model than supervised training.
Our \methodname framework outperforms the existing self-training methods consistently across different datasets and annotation settings.

\subsubsection*{Generality across different EA models}

To check whether these self-training methods work well for different EA models (e.g. in terms of model architecture and performance), we run them with another three EA models: RREA~\cite{DBLP:conf/cikm/MaoWXWL20}, AliNet~\cite{DBLP:conf/aaai/SunW0CDZQ20}, and GCN-Align~\cite{DBLP:conf/emnlp/WangLLZ18}.
Fig.~\ref{fig:generality_neural_models} reports their results on the zh\_en dataset (results on the other datasets show a quite similar trend). 
We have the following findings:
(1) The choice of EA model can affect the impact of self-training methods. For example, the existing three self-training methods have slighter impact on AliNet than on the other EA models.
(2) Our \methodname framework can always outperform the existing self-training methods when running with the different EA models under different datasets and annotation settings.
(3) In few annotation settings (e.g. 1\% and 5\%), the advantage of \methodname over the baselines is more obvious than that in rich annotation settings. This is because the baselines depend on the model's confidence and suffer more from the introduced noise when the EA model is poorly trained.
(4) In terms of the final EA effectiveness, the SOTA EA models (i.e. Dual-AMN and RREA) in supervised mode can achieve the best performance in self-training mode as well.

\noindent
In short, the superiority of \semiea over the baselines is generic across different EA models.

\subsubsection*{Reliance on data annotation}

\begin{figure}
    \centering
    \includegraphics[width=8.8cm]{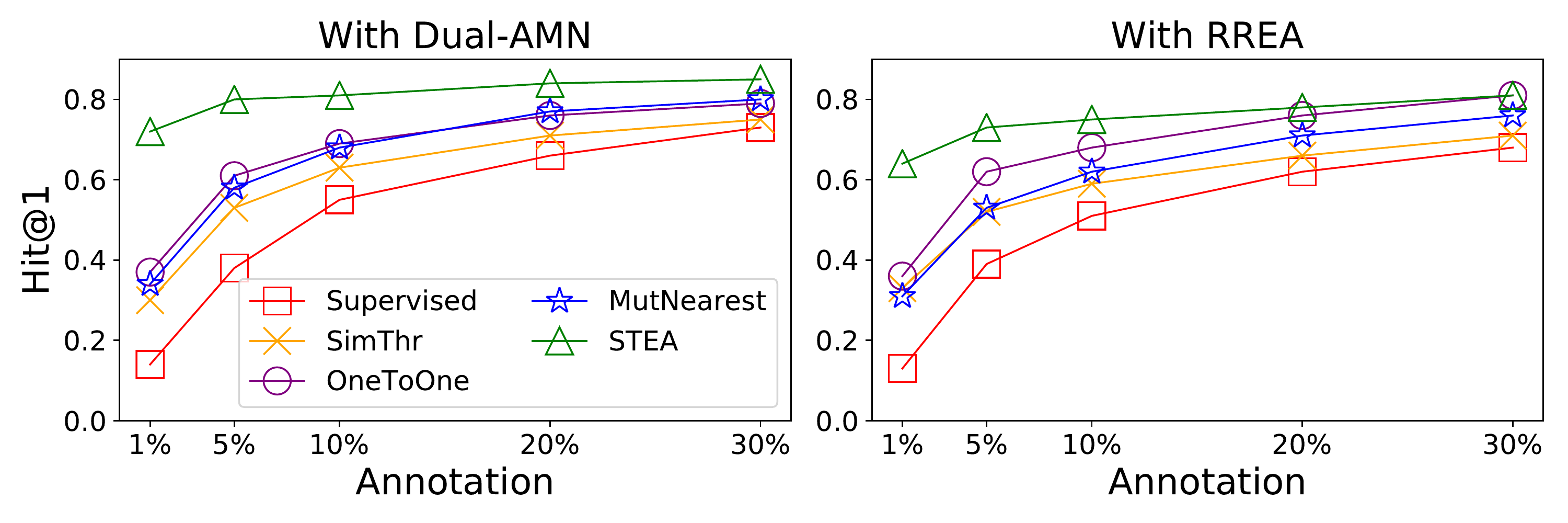}
    \vspace{-20pt}
    \caption{Performance change of self-training methods w.r.t. annotation amounts. The \methodname can achieve decent effectiveness in extremely few annotation scenario, while sharp increasement in annotation cannot bring quick improvement.}
    \vspace{-12pt}
    \label{fig:reliance_on_anno}
\end{figure}

As known, the motivation of self-training is to alleviate the models' reliance on annotations.
Therefore, we specially examine the effect of annotation amount on the performances of the self-training methods with Dual-AMN and RREA (two SOTA EA models) on dataset zh\_en.
In Fig.~\ref{fig:reliance_on_anno}, each curve depicts the performance change of a certain self-training method w.r.t. the increasing amount of annotations. 
We can observe that: 
(1) Even though the three baselines can always boost supervised training, their performances are still poor when the training data is few;
(2) On the contrary, \methodname can achieve pretty decent effectiveness. With only 1\% labelled data, \semiea can achieve comparable performance as supervised learning using 30\% labelled data. Similar observations on the other datasets can be obtained from Table~\ref{tab:overall_perf}. 
From the perspective of self-training for EA, this finding is inspiring. The value of self-training for EA is much beyond what has been realised. Thus, this direction deserves more exploration.
(3) The performance of \methodname is relatively stable, like reaching a ceiling, even the annotation amount is increased sharply from 1\% to 30\%. We reckon this phenomenon indicates that random data selection for annotation is not suitable for the self-training based EA methods. Active Learning is a promising direction of smart data selection. Though it has been explored by a few recent works for EA task~\cite{DBLP:conf/emnlp/LiuSZHZ21}, the combination of active learning with self-training remains to be investigated.

\noindent
In conclusion, \methodname can greatly alleviate the reliance of EA models on annotation. Facing the performance ceiling, active learning for self-traininng is a promising direction to be studied.

\subsubsection*{Precision and recall of pseudo mappings.}
% precision, recall.
 
\begin{figure}
    \centering
    \includegraphics[width=8.7cm]{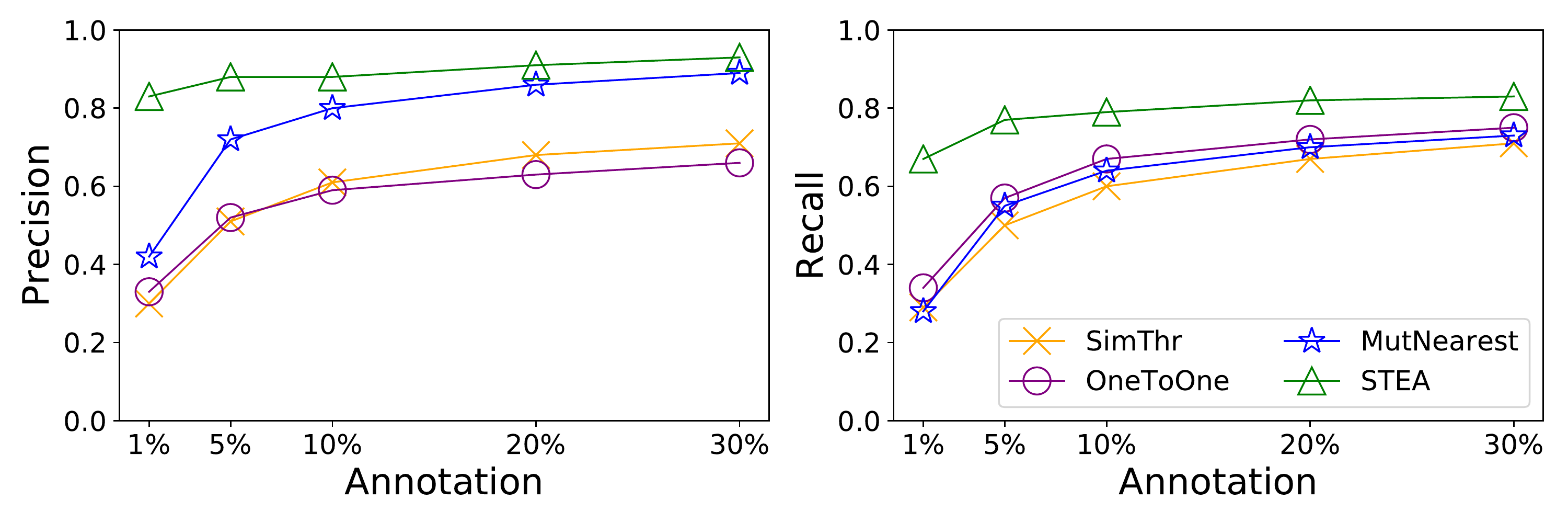}
    \vspace{-20pt}
    \caption{Precision and recall of pseudo mappings  w.r.t. annotation amount. \methodname has obvious advantage in both precision and recall regardless of the annotation amount.}
    \vspace{-12pt}
    \label{fig:prec_recall_annotation}
\end{figure}

\begin{figure}
    \centering
    \includegraphics[width=8.7cm]{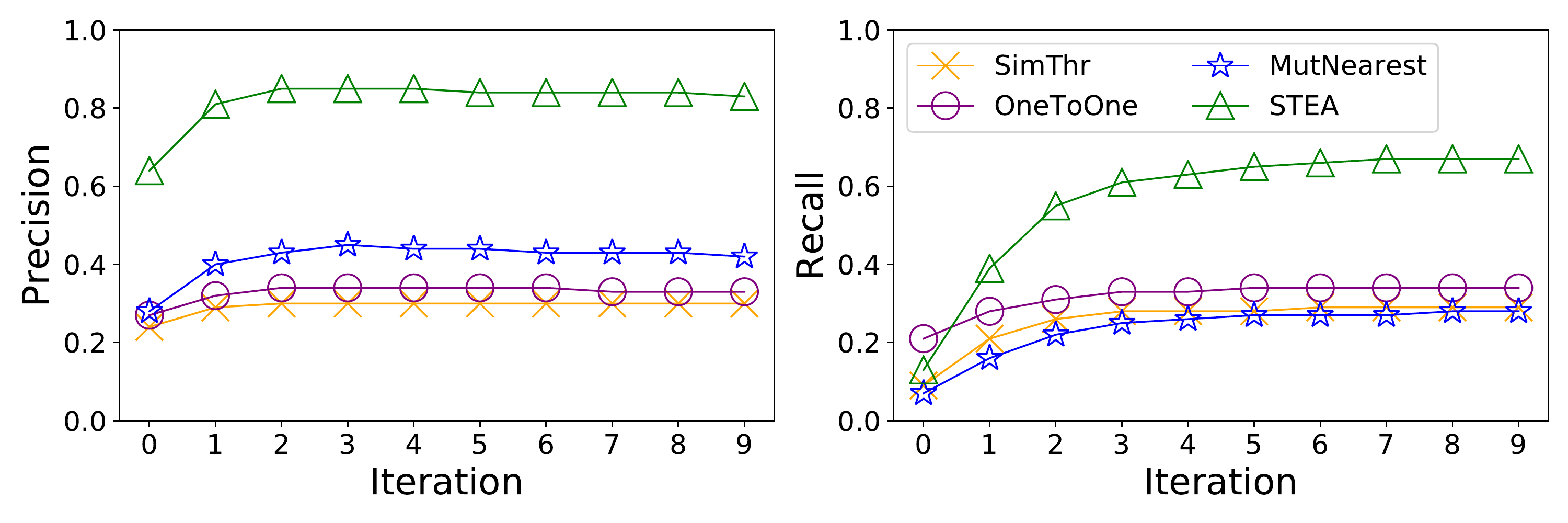}
    \vspace{-20pt}
    \caption{Precision and recall of generated pseudo mappings in each iteration when the annotation amount is 1\%. \methodname is able to suppress the noise without sacrificing recall from the beginning iteration.}
    \label{fig:prec_recall_iteration}
    \vspace{-10pt}
\end{figure}

High precision means there is less noise, while high recall of ground-truth mappings means more true training data are added.
To get more insight into the advantage of \methodname, we check these two metrics for different self-training methods.
Fig.~\ref{fig:prec_recall_annotation} shows these metrics of finally generated pseudo mappings by different self-training methods w.r.t. different annotation amounts on zh\_en.
We can observe that \methodname has a big advantage in both precision and recall over the baselines regardless of the amount of annotation. It is impressive to see \methodname still can achieve high precision in few annotation settings thanks to its dependency-aware predictions.
Furthermore, we take a closer look at the generated pseudo mappings in each iteration when the annotation amount is 1\%.
As shown in Fig.~\ref{fig:prec_recall_iteration}, \methodname has high precision from the beginning (when the EA model has poor effectiveness), while its recall increases gradually from a low level.
On the contrary, the baselines can introduce much noise in the beginning and then can hardly get improved in the later iterations.

\noindent
Thus, we reckon the success of \methodname comes from: the exploitation of dependency can help suppress the noise without sacrificing high recall of ground-truth mappings.

\vspace{-4pt}
\subsection{Detailed Anlysis of the \methodname}

% \subsubsection*{Why \methodname}
% * Quality of pseudo-mappings.
% * Error propagation. error exaggeration.

\subsubsection*{Comparing strategies of selecting pseudo mappings within \methodname}

\begin{figure}
    \centering
    \includegraphics[width=8.7cm]{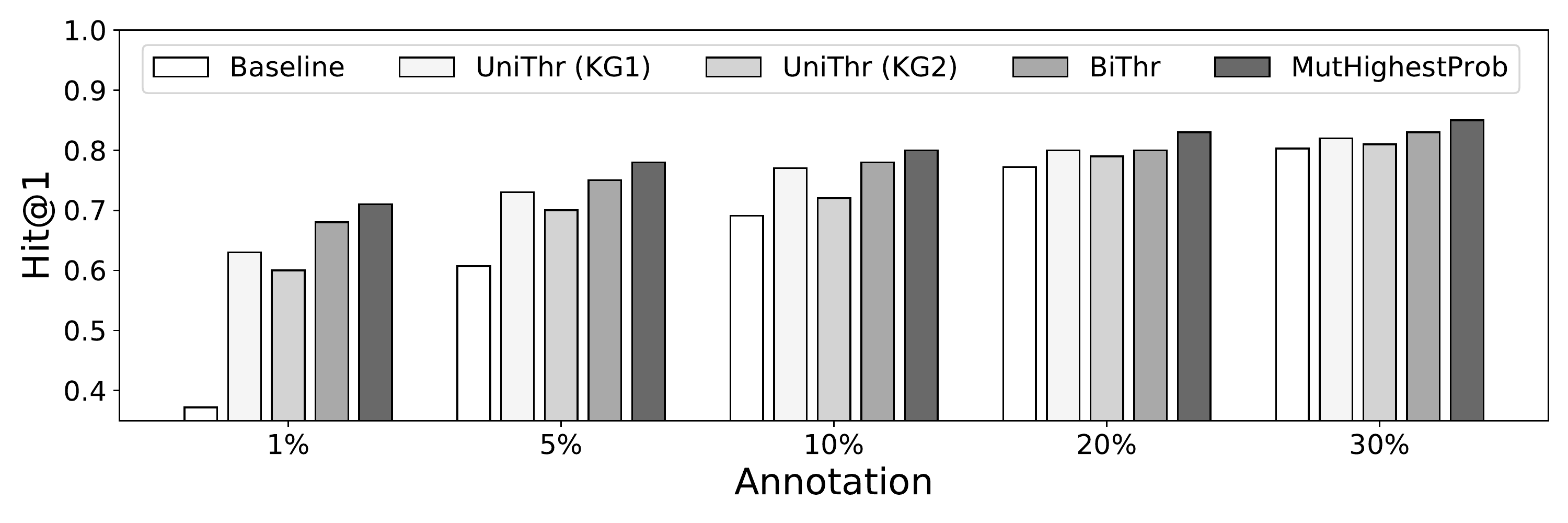}
    \vspace{-20pt}
    \caption{Comparing different strategies of deriving pseudo mappings in \methodname. MutHighestProb can achieve the best performance and is also easiest to use in practice.}
    \vspace{-12pt}
    \label{fig:uni_bi_mut}
\end{figure}

Based on the dependency-aware probabilities, we explore three types of strategies to generate the pseudo mappings -- UniThr, BiThr, MutHighestProb. Among them, UniThr has two variants which do compatibility checking only from the first KG (denoted as \textit{UniThr (KG1)}) or only from the second KG (denoted as \textit{UniThr (KG2)}).
In addition, we use \textit{Baseline} to denote the best baseline within each annotation setting.
Fig.~\ref{fig:uni_bi_mut} shows their performances on the zh\_en dataset. As can be seen:
(1) The choice of KG for UniThr may have a big effect on the final performance.
(2) BiThr always works better than UniThr (both variants). Meanwhile, it does not need to choose source KG and thus only need to adjust one hyperparameter -- threshold.
(3) MutHighestProb can achieve the best performance. In addition, it is the easiest strategy to use in practice since it has no hyperparameter.
(4) In most cases, these strategies outperform the best baseline.

\subsubsection*{Necessity of the normalisation module.}
To normalise the similarities into probabilities, we use one learnable normalisation module based on \textit{softmax} instead of a simple one, e.g. MinMax scaler. To verify the necessity, we replace our normalisation module with a MinMax scaler in \methodname.
Table~\ref{tab:normalisation} shows the results on dataset zh\_en. Obviously, MinMax scaler performs poorly. We think the reason is: the normalised similarity vector by MinMax scaler only contains very small values and the highest value is not distinguishable from the others. Thus, it is not suitable to be used as probability distribution. 

\subsubsection*{Sensitivity to hyperparameter}
\begin{figure}
    \begin{minipage}{0.25\textwidth}
        \includegraphics[width=4.3cm]{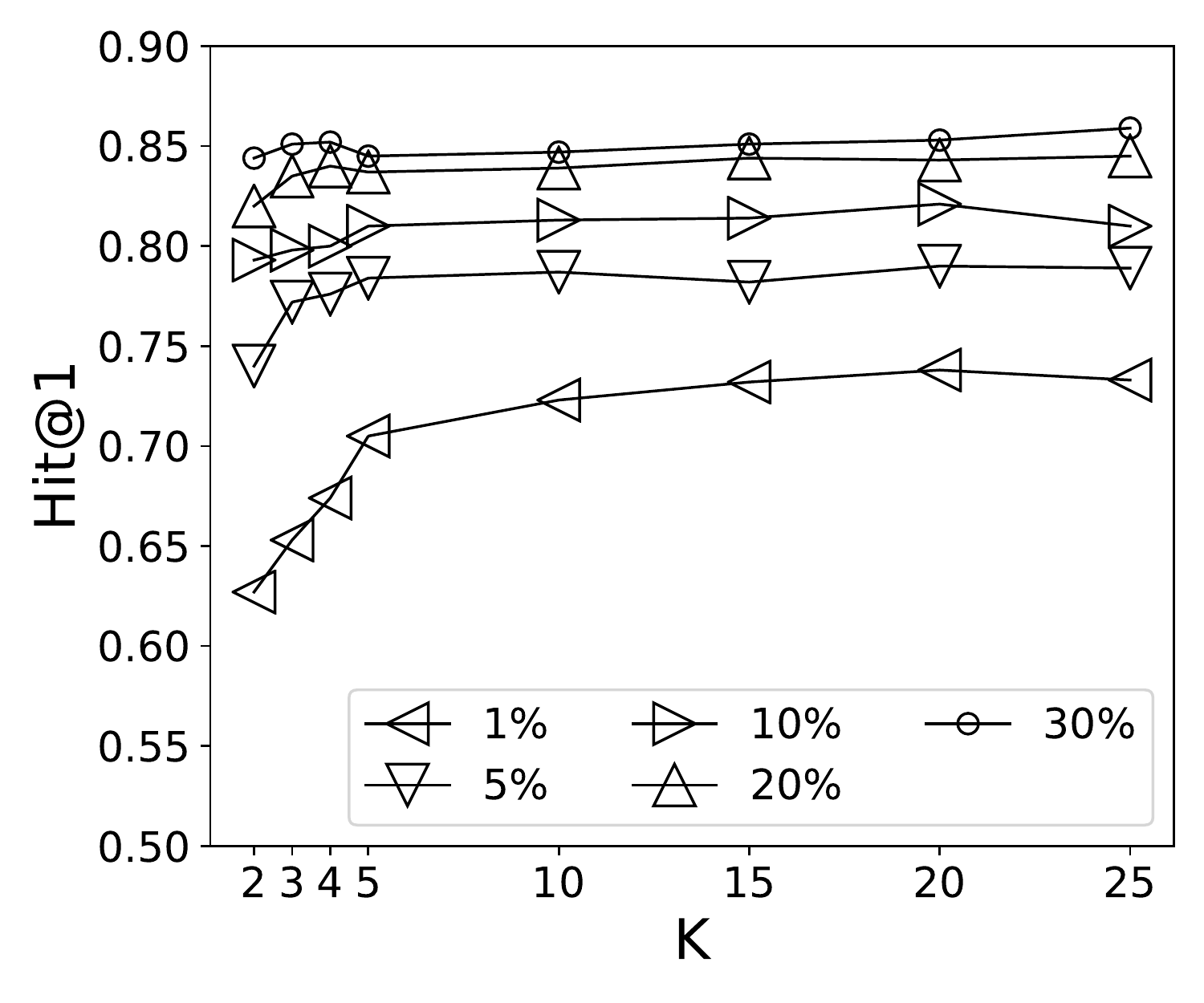}
        \vspace{-16pt}
        \captionof{figure}{Performance of \methodname w.r.t. hyperparameter $K$ under different annotations. \methodname is not sensitive to $K$.}
        \vspace{-12pt}
        \label{fig:paramK}
    \end{minipage}%
    \hspace{2mm}
    \begin{minipage}{0.2\textwidth}
        % \begin{table}
            \captionof{table}{Comparison of normalisation methods in \methodname. Simple MinMax Scaler leads to very poor effectiveness (Hit@1).}
            \label{tab:normalisation}
            \scalebox{0.85}[0.85]{
                \centering
            \begin{tabular}{|c|c|c|}
                \hline
                Anno. & MinMax & Softmax \\
                \hline
                1\% & 0.48&0.59\\
                5\% & 0.59&0.74\\
                10\% & 0.65&0.79\\
                20\% & 0.72&0.83\\
                30\% & 0.74&0.84\\
                \hline
            \end{tabular}
            }
    \end{minipage}
\end{figure}

\methodname only has one hyperparameter $K$, which means, for each source entity, only the top $K$ counterpart candidates suggested by the neural EA model will be estimated for compatibility to simplify the computation.
In Fig.~\ref{fig:paramK}, we show the performance of \methodname w.r.t. different $K$ values under different annotation settings. 
(1) \methodname is only sensitive to very small values of $K$ ($<5\%$). 
(2) Fewer annotations make \methodname more sensitive to $K$ with small values ($<5\%$).
(3) We prefer to set $K$ as a small value out of the sensitive interval like 10, considering larger $K$ will lead to higher computation cost.

\section{Conclusion}

Entity Alignment is a primary step of fusing different KGs. Though neural EA models have achieved promising performance, their reliance on labelled data is still an open problem.
To address this issue, a few self-training strategies have been explored and shown effective in boosting the training of EA models.
However, self-training for EA is never studied systematically. 
In addition, the self-training strategies used in existing works have limited impact because of the introduced noise.
In this work, we expand the knowledge about self-training for EA by benchmarking the existing self-training strategies and evaluate them in comparable experimental settings.
Furthermore, towards more effective self-training strategy for EA, we propose a new self-training framework named \methodname. This framework features in exploiting the dependencies between entities to detect suspicious mappings and improve EA predictions. Based on the derived dependency-aware EA predictions, we further explored different ways for generating pseudo mappings to be included in self-training.
We empirically show that \methodname outperforms the existing methods with a big margin across different datasets, annotation amounts, and neural EA models. 
\methodname can greatly reduce the reliance of EA on annotations. In particular, \methodname using 1\% labelled data can achieve decent effectiveness, which is equivalent to supervised training using 30\% labelled data. 

In future, we plan to investigate the combination of active learning and self-training for EA. As shown, in the self-training based EA, random data selection is very inefficient in improving the EA performance. With active learning, we aim to reach the desired EA performance with the least annotation effort.

\begin{acks}
  This research is supported by the National Key Research and Development Program of China No. 2020AAA0109400, the Shenyang Science and Technology Plan Fund (No. 21-102-0-09), and the Australian Research Council (No. DE210100160 and DP200103650).
\end{acks}

\bibliographystyle{ACM-Reference-Format}
\bibliography{bibfile}

% \appendix

% \input{sections/method_math.tex}

\end{document}